6th International Building Physics Conference, IBPC 2015

# Towards a general framework for an observation and knowledge based model of occupant behaviour in office buildings

Khadija Tijani[1,2,3]\*, Quoc Dung Ngo[4], Stephane Ploix[1], Benjamin Haas[2], Julie Dugdale[3]

[1]*Univ. Grenoble Alpes, G-SCOP, F- 38000Grenoble, France*
[2]*Centre Scientifique et Technique du Bâtiment, CSTB, France*
[3]*Univ. Grenoble Alpes, LIG, F-38000 Grenoble, France & Univ. Agder, 9 4879 Grimstad, Norway*
[4]*Post and Telecommunication Institut of Technology, Hanoi, Vient Nam*

**Abstract**

This paper proposes a new general approach based on Bayesian networks to model the human behaviour. This approach represents human behaviour withprobabilistic cause-effect relations based not only on previous works, but also with conditional probabilities coming either from expert knowledge or deduced from observations. The approach has been used in the co-simulation of building physics and human behaviour in order to assess the $CO_2$ concentration in an office.
© 2015 The Authors. Published by Elsevier Ltd.
Peer-review under responsibility of the CENTRO CONGRESSI INTERNAZIONALE SRL.

*Keywords:*occupant behaviour, indoor air quality, bayesian network

## 1. Introduction

Most of the world standards for buildings take occupants into account by representative values and deterministic scenarios: number of occupants, predefinedschedules, and responses to an exceeded threshold. Nevertheless, in low consumption buildings, the impact of occupant behaviour stronglyinfluences energy consumption and indoor climate conditions. For instance, if air quality is poor, occupants may decide to open windows or doors, which may depend on other occupants' activities or wishes. Thereforeit is important to take into account occupant behaviourwhen estimating energy consumption.

A high performance building should be efficient for a large diversity of use. Therefore, in building simulation it is necessary to simulate the physical aspects of the building with its appliances, and the behaviour of its reactive occupants i.e. thoseoccupant behaviours affectingindoor conditions such temperature or air quality. Numerical models usually take into account occupant behaviour based on presence and ondifferent profiles foropening windows.However, this may be insufficient since some simulation software handles the occupant model and the windows model separately, potentially leading to a situation where the occupant is absent, but the window is opened. In addition, profiles rely on datacollected from specific observations. This makes it difficult to adapt them to otherbuildings or homes. Thispaper proposes a design methodology based on a knowledge model of occupant behavioursimilar to the models used in building physics. Occupants' behavioursare designed using an a priori knowledge of future occupants and observations of behaviour to tune some of the model's parameters. This model is then used for co-simulating physical and human aspects. The approach relies on Bayesian Networks (BN) and represents human behaviour by probabilistic cause-effect relations based on experts' knowledge, andconditional probabilities coming either from the latteranalysed observations.

\* Corresponding author. Tel.: 0033634288214.
 *E-mail address:* khadijatijani@gmail.com

The approach is applied to the co-simulation of an office combining a behavioural model representing occupancy and actions on doors and windows, and a $CO_2$ physical models. The modelof the human behaviour is tuned according to experimental data.

## 2. State of the art

In the scientific community of building physics, simulation programs are becoming more advanced and there has been much effort to improve the prediction accuracy by taking in account occupant behaviour. Studies have shown its importance for energy waste reduction in buildings [1]. Lee and his colleagueshave explained the relation between behaviour and energy performance by using Energy Plus for simulating dynamic occupant behaviour [2].

In the literature, three kinds of approaches can be found: deterministic approaches based on predefined scenarios and behavioural rules;statistical approachesthat rely on factual observations such as surveys; and social modelling approachesthat model cognitive and deliberative behaviours.In a deterministic approach, behavioural rules such as"if the temperature exceeds 28°C then open the window" may be designed. However some researchers believe that human behaviour is better represented by stochastic models [3].[4] defend the ideathat occupant behaviour corresponds to stochastic processes and not to deterministic ones: "there is no precise temperature at which everyone will open a window, but the higher the temperature, the higher the probability of opening the window". [5] studies the probability of opening and closing windowsas afunction of the indoor and the outdoor temperatures.The authors established a probability distribution of actions as afunction of the indoor temperature. In 2009, inspired by previous works, [3] proposed a hybrid stochastic model for window opening based on three modelling approaches: logistic probability distributions, Markov chains and continuous-time random processes.

More recently, a new approachto social modelling has been proposed that is based on agents. Here computational agents, each having their own characteristics and behaviours, are used to model the occupants.[6] uses this approach to describe occupants' behaviourfor energymanagement. The BRAHMS (Business Redesign Agent-Based Holistic Modelling System)platform was used to model each agent's cognitive behaviour. BRAHMS is based on activity theory and is able to model an occupant's (agent's) beliefs, desires and intentions (BDI). The BDI architecture allows occupant's cognitive, reactive and deliberative behaviours to be represented. [8] uses neural networks for the agents to learn their behaviours from recorded data to ensure their comfort. After a learning phase, agents know the actions that increase their comfort in different environmental conditions. In a recent study, [8] models various occupancy profiles by calibration to predict the use of fans, heating and windows.

The social modelling approach is able to capture the same level of complexity as Markov chain processes. Combined with field studies, the agent-based approach proposes cognitive and deliberative schemas that go beyond statistical approaches. However, the complexity of social modelling makes it difficult to be used by designers[9]: a more focused modelling approach, described in the following section, would be more helpful.

## 3. Using Bayesian Networks as a general tool for human behaviour modelling

A Bayesian Network (BN) is a probabilistic causal model represented by a directed acyclic graph. Intuitively, it is both a knowledge model and an inference engine using conditional probabilities and evidence to deduce resulting probabilities. Causal relationships between variables are assigned probabilitiesand are represented inagraph. The observation of a variable does not automatically invoke the related effects but changes the probability of observing them. Because of the causal representation, a BN is a human friendly way to represent complex behaviours that can be summarized by conditional probabilities.AStatic BN (SBN) only concerns a single slice of time and is not appropriate for analyzing a system that changes over the time. ADynamic BN (DBN) overcomes this problem by describing how variables influence each other over time based on a model derived from past data. A DBN represents the relations between variables at different time steps, some of these variables may not be directly observed.

Compared to a BN, Hidden Markov Models (HMM) rely on nodes that correspond to exclusive states.Nevertheless,the possible number of states for human behaviour isnumerous. A BN is more general that HMM because each node stands for a variable (that could possibly be a state) corresponding to any fact, feeling, belief, desire orintention i.e. the basic concepts in social modelling.From a more technical point of view, [10] states that the accuracy of inferences done by DBNs is less sensitive than HMM using the same amount of observations. Unlike HMM, BNs offer an easy way to combine prior expert knowledge with conditional probabilities learnt from observations. This point significantly enhances the precision ofuser behaviour modelling because merely learning a statistic model from data is generally not rich enough for user activities recognition.

To build a BN, a causal graph andconditional probability tables have to be defined for each node and refined usingexpert knowledge or from observations. There are two approaches to designthe structure of the causal graph.The first approachconsists in learning the causality directly from data, if they exist;howeverdue to poor data, the resulting causalities are often inconsistent with reality. The second approach, and the one used in this work, is based on designinga causal graph using expert knowledge about occupant cognition in different contexts and activities.

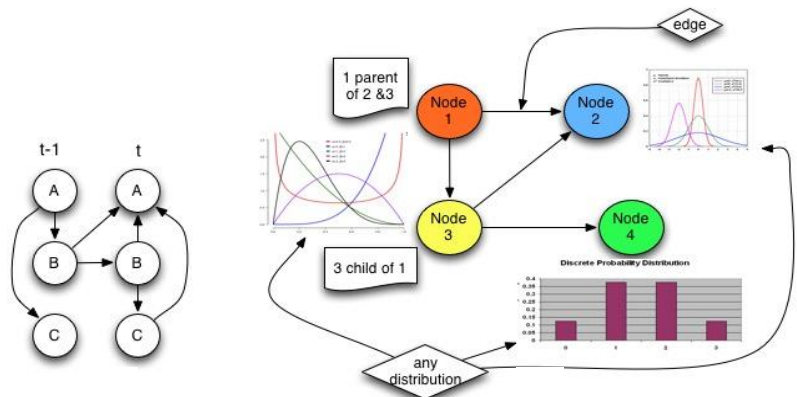

Fig.1 Elements of a Bayesian Network: (a) Dynamic BN (b) Different probability distributions

In BN modelling, anytype of probability distribution function (PDF), e.g. uniform, normal, Poisson PDF (fig.1(b)), can be used with a directed edge between 2 nodes when discretized domain values exist for the input variable. However, nonlinear functions of variables cannot be represented. This workproposes the combination ofa DBN with nonlinear functions, for instance,Fanger'sPMV model, to process output inferences.

## 4. Co-simulation mechanism

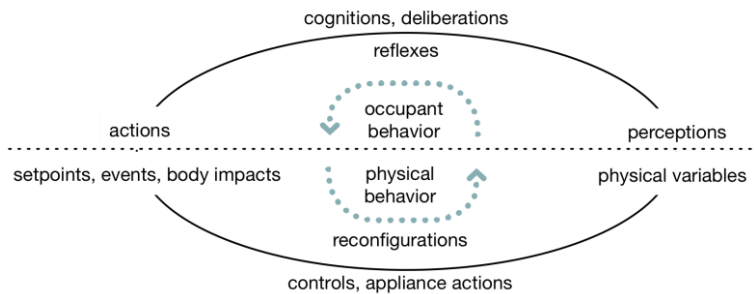

Fig.2 Elements of the model

According to [7], occupants control comfort via a loop summarized in fig.2. Therefore, the simulation is designed as a loop where the physical part (below the dotted line) plays the role of an orchestrator, even if the occupant part (above the dotted line) or an external component could also play this role. The occupant part contains step methods, wich is a method that manage the question/response between the physical part and occupant one. Because of the time step, actions and perceptions are averaged. For instance, instead of considering the 2 natural states of 'opened' or 'closed' for a door, different states such as 'always closed', 'mostly closed', 'mostly opened' and 'always opened' are used. For example, if the tipme step is one hour, some times the door is not closed for the hole hour, it may be close just 30% of the time step, for this case, it should be provided another door state like 'mostly opened'.

BNs are general tools for modelling human behaviour where the graph nodes could represent any element of a cognitive and deliberative model such those proposed in [6]. These include perceptions, feelings, beliefs, desires depending on cognition and deliberations, intentions and actions. The occupant behaviour simulator is composed of a DBN editor, that uses the 'libpgm' library adapted to Python 3, and a library of existing DBNs representing elements of occupant behavioural models. It is thus possible to build the behaviour of occupants just as a home setting is built, taking into account the available knowledge about occupants and some assumptions, or reusing existing models. A composition tool is used to combine model elements. A scenario editor can generate inference for different times and dates. Databases of scenarios and results, as well as a result viewer, are available.

The proposed approach couples occupant behaviour with air quality energy performance calculation tools, which rely on dynamic energy calculation rather than on pure thermal calculation. Indeed, without energy calculations that take into account HVAC systems, the coupling with the occupant is less informative. The work is based on an existing tool from CSTB (Scientific and Technical Centre for Building, a French public organisation) named COMETh (COre for Modelling Energy consumption and THermal Comfort) that already includes models for a wide range of HVAC systems (mostly those modeled in the EPBD CEN set of standards). Coupling occupant behaviour with physical aspects is implemented by overriding the control modules with a consistent occupant module.

Specific critical aspects related to occupants actions and air quality must be defined. Because the occupant models are coupled with physical models, the occupants' actions that must be modelled are constrained by the physical computations. These values are: a Boolean indicating whether the space is occupied or not; metabolic gains per thermal zone; activated appliances with corresponding gains per thermal zone; metabolic humidity gains per thermal zone; set point temperature; heating and cooling period if managed by the occupant; ratio of window opening; ratio of door opening. An occupant model must provide all of these values. Because actions and perceptions are averaged, transient states must be carefully designed. In addition, the physical simulation must provide the values that allow the occupant to make decisions and take actions. A minimal set of data, that can enrich the simulation, is the operative mean temperature, mean humidity and $CO_2$ concentration.

## 5. Application to door management and air quality in an office

The chosen office is non-mechanically ventilated and has sensors to capture data on: temperature, presence, $CO_2$ levels, door and window contacts, and humidity. Video recording equipment is also used to track occupants movements and actions. The office can accommodate up to 4 persons, but in order to show the approach only one occupantand a visitor are modelled. The following actions of an office occupant have been derived from analysing video data. The occupant, a Professor, is not always in the office, but when he is, he works on his computer and usually remains seated. He often closes the door for privacy andwelcomes visitorsin his office.The DBN structure is shown in fig.3. The target node "Door Movement" is influenced by six variables, identified through video data, and represented by nodes. Each node has its own conditional probabilities. The domain values of the variables are represented by blue rectangles. The structure is dynamic because it uses the "past door movement" variable.

The aim of this model is to predict the rate of $CO_2$ concentrations in the office depending on the door opening. To design the model, sensor data were collected, and physical models of $CO_2$ generation [11] and flow calculations [12] were used. The equation (1) is based on mass balance considerations in a zone with a flow of air renewal.

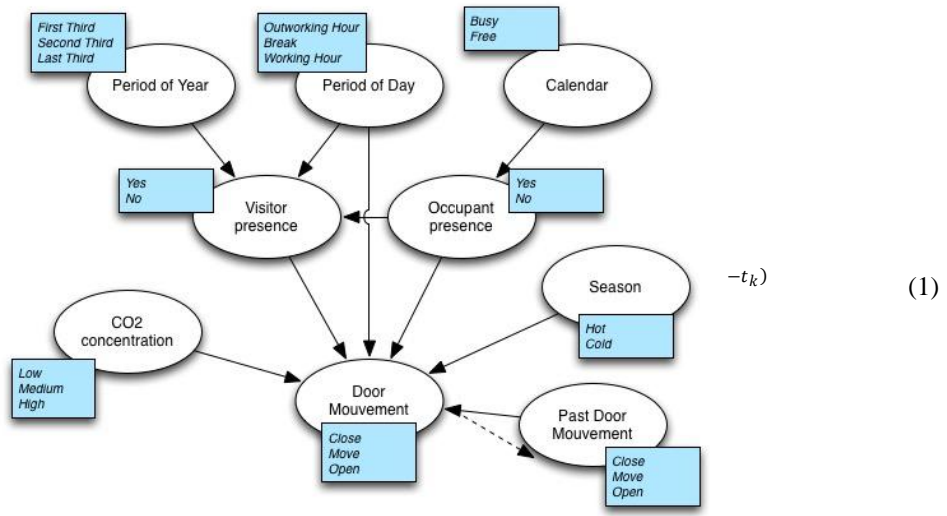

$$\qquad\qquad\qquad\qquad\qquad\qquad\qquad\qquad\qquad\qquad -t_k) \qquad (1)$$

Fig.3 Element of the DBN structure

where $C(t)$ is the $CO_2$ concentration inside the building at time $t$, $C_i$ is the initial concentration of $CO_2$, $V$ is the office volume, $S$ is the generation rate of $CO_2$ from occupants exhaled air,which is a function of the volumetric rate of $CO_2$ [13]. $Q_i$ and $Q_o$ are the incoming or outgoing volumetric flow rate of renewed air. They aredefined by equation(2) which calculates the airflow passing through a large opening between the corridor and the office.

$$Q_{i,o}(z_1, z_2) = \frac{2}{3\rho}\epsilon CL\sqrt{2\rho|\Delta\rho|g}\left(|HN - z_1|^{3/2} - |HN - z_2|^{3/2}\right) \qquad (2)$$

where $Q_{i,o}$ is the incoming or outgoing air flow, $\epsilon = \Delta T/|\Delta T|$, $\rho$ is the air density (dependent on the corridor and office air densities), $g$ is gravity, $HN$ neutral plan (when the pressure difference between two areas is 0), $z_{1,2}$ are the heights of the opening that can be $HN$, low or high height depending on the temperatures.

To illustrate the approach, two working calendar days,in the first third of the year (months 1-4), of the occupant's activity are considered (fig.4).

| Hour | 0 | 1 | 2 | 3 | 4 | 5 | 6 | 7 | 8 | 9 | 10 | 11 | 12 | 13 | 14 | 15 | 16 | 17 | 18 | 19 | 20 | 21 | 22 | 23 |
|---|---|---|---|---|---|---|---|---|---|---|---|---|---|---|---|---|---|---|---|---|---|---|---|---|
| Working day calendar 1 | Out of working time | | | | | | | free | free | busy | busy | free | Lunch | | busy | busy | busy | free | free | free | free | Out of working time | | |
| Working day calendar 2 | Out of working time | | | | | | | busy | busy | busy | busy | busy | Lunch | | busy | busy | busy | free | free | busy | busy | Out of working time | | |

Fig.4 Activities for two calendar working days

Each working day calendarhas been generated by co-simulation100 times yieldingthe results in fig.5. These resultsshow that the door is less used when the occupant is busy and even less during lunch, meaning that the occupant has lunch outside. This can be seen by comparing the two free times shown in green in fig.5(b) where at 5pm the door is used more often than during the lunch break. Fig.5clearly shows that the proposed DBN can handle the differences between two same inputs (here free) but with different activities (lunch break and free for some reason).The results are different and illustrate the real aim of the activity.

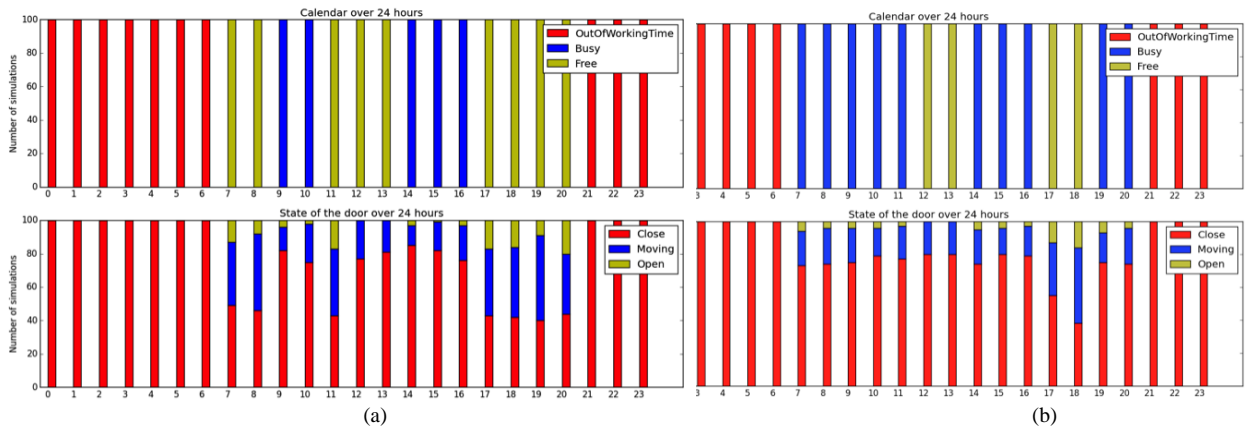

(a) (b)

Fig.5 Simulations with the (a) first (b) second working day calendar

## 6. Conclusion

A new approach based on a hybrid DBNhas been proposed to model human behaviour. Causal representation eases the modelling process, asit is a natural way of thinking about human behaviour. Furthermore, a probabilistic modelling approach provides a relevant level of representing human behaviour, which becomes very complex at a finer level. Human behaviour can then be designed step-by-step, just like the physical part of a building system, taking into account the knowledge about future occupants and possible assumptions about their behaviours. This approach has the advantage of being intuitive, like cause-effect human reasoning. It also takes into account both the knowledge model and the probability distribution functions, which is missing in other approaches.

This approach has been illustrated in the co-simulation of $CO_2$ concentration with human behaviour in an office. It will be extended by adding more occupants and performing co-simulations with dynamic energy simulation software.